\documentclass{article}

\usepackage[preprint]{neurips_2023}

\usepackage[utf8]{inputenc} 
\usepackage[T1]{fontenc}    
\usepackage{hyperref}       
\usepackage{url}            
\usepackage{booktabs}       
\usepackage{amsfonts}       
\usepackage{nicefrac}       
\usepackage{microtype}      
\usepackage{xcolor}         

\usepackage{bm}
\usepackage{amsmath}
\usepackage{algorithmicx}
\usepackage{algorithm,algpseudocode}
\usepackage{graphicx}
\usepackage{multirow}
\usepackage[small]{caption}
\usepackage[title]{appendix}
\usepackage{subfigure}
\usepackage{multicol}

\title{Learning from Long-Tailed Noisy Data with

Sample Selection and Balanced Loss}

\author{%
  Lefan Zhang, Zhang-Hao Tian, Wujun Zhou, Wei Wang\thanks{Corresponding author}\\
  National Key Laboratory for Novel Software Technology\\ 
  Nanjing University, Nanjing 210023, China \\
  \texttt{\{zhanglf, tianzh, zhouwujun, wangw\}@lamda.nju.edu.cn}
}

\begin{document}

\maketitle


\begin{abstract}
The success of deep learning depends on large-scale and well-curated training data, while data in real-world applications are commonly long-tailed and noisy. Many methods have been proposed to deal with long-tailed data or noisy data, while a few methods are developed to tackle long-tailed noisy data. To solve this, we propose a robust method for learning from long-tailed noisy data with sample selection and balanced loss. Specifically, we separate the noisy training data into clean labeled set and unlabeled set with sample selection, and train the deep neural network in a semi-supervised manner with a balanced loss based on model bias. Extensive experiments on benchmarks demonstrate that our method outperforms existing state-of-the-art methods.
\end{abstract}

\vspace{-0.3cm}
\section{Introduction}
\label{introduction}
\vspace{-0.1cm}

Deep neural networks have made great successes in machine learning applications \citep{he2016deep,vaswani2017attention} but require well-curated data for training. These data, such as ImageNet \citep{russakovsky2015imagenet} and MS-COCO \citep{lin2014microsoft}, are usually artificially balanced across classes with clean labels obtained by manual labeling, which is costly and time-consuming. However, the data in real-world applications are long-tailed and noisy, since data from specific classes are difficult to acquire and labels are usually collected without expert annotations. To take WebVision dataset as an example, it exhibits long-tailed distribution, where the sample size of each class varies from 362 (Windsor tie) to 11,129 (ashcan), and contains about 20\% noisy labels \citep{Li2017a}. Thus, developing robust learning methods for long-tailed noisy data is a great challenge.

Many methods have been proposed for long-tailed learning or learning with noisy labels. 
In terms of long-tailed learning, re-sampling methods \citep{chawla2002os1,jeatrakul2010us1}, re-weighting methods \citep{cui2019cbloss,cao2019ldam,menon2020long}, transfer learning methods \citep{liu2019large,Kim2020m2m} and two-stage methods \citep{kang2019decoupling,cao2019ldam} are included; 
in terms of learning with noisy labels, designing noise-robust loss functions \citep{ghosh2017robust,zhang2018generalized}, constructing unbiased loss terms with the transition matrix \citep{patrini2017making,hendrycks2018using}, filtering clean samples based on small-loss criterion \citep{han2018co,li2020dividemix} and correcting the noisy labels \citep{tanaka2018joint,yi2019probabilistic} are included. 
Despite learning from long-tailed or noisy data has been well studied, these methods cannot tackle long-tailed noisy data in real-world applications. A few methods are proposed to deal with long-tailed noisy data, which mainly focus on learning a weighting function in a meta-learning manner \citep{shu2019meta,jiang2021delving}. However, these methods simultaneously require additional unbiased data which may be inaccessible in practice. 


To deal with long-tailed noisy data, an intuitive way is to select clean samples with small-loss criterion and then apply long-tailed learning methods. For the sample selection process, \citet{gui2021towards} revealed that the losses of samples with different labels are incomparable and chose each class a threshold for applying the small-loss criterion. For long-tailed learning, existing methods are commonly based on label frequency to prevent head classes from dominating the training process. However, the model bias on different classes may not be directly related to label frequency (see Appendix \ref{appendix:frequency}), and the true label frequency is also unknown under label noise. In this paper, we propose a robust method for learning from long-tailed noisy data. Specifically, we separate the noisy training data into clean labeled set and unlabeled set with class-aware sample selection and then train the model with a balanced loss based on model bias in a semi-supervised manner.
Experiments on the long-tailed versions of CIFAR-10 and CIFAR-100 with synthetic noise and the long-tailed versions of mini-ImageNet-Red, Clothing1M, Food-101N, Animal-10N and WebVision with real-world noise demonstrate the superiority of our method.

\vspace{-0.3cm}

\section{Related Work}

\vspace{-0.2cm}

\textbf{Learning from Long-Tailed Data.}
The purpose of long-tailed learning is to alleviate the performance degradation caused by the small sample sizes of tail classes. Popular methods include re-sampling/re-weighting methods which commonly simulate a balanced training set by paying more attention to tail classes \citep{jeatrakul2010us1,lin2017focal,cui2019cbloss}, transfer learning methods which boost the recognition performance on tail classes by utilizing the knowledge learned from head classes \citep{liu2019large,Kim2020m2m} and two-stage methods which argue that decoupling the representation learning and the classifier learning leads to better performance \citep{kang2019decoupling,cao2019ldam}. Recently, a new line of works are proposed, which utilize the contrastive learning to learn a class-balanced feature space and further achieve better long-tailed learning performance \citep{yang2020rethinking,kang2021exploring}. 

\noindent
\textbf{Learning from Noisy Data.}
Works proposed to address the problem of learning with noisy labels can be divided into three categories: 1) methods based on robust loss \citep{ghosh2017robust,zhang2018generalized}, 2) methods based on loss correction \citep{patrini2017making,hendrycks2018using} and 3) methods based on noise cleansing \citep{han2018co,tanaka2018joint,yi2019probabilistic}. According to whether remove or correct the noisy labels, the third category can be further divided into sample selection methods and label correction methods. Sample selection methods try to identify clean samples, which are commonly based on small-loss criterion \citep{han2018co,gui2021towards}. Label correction methods try to improve the quality of raw labels \citep{tanaka2018joint,yi2019probabilistic}. Sample selection can also be combined with label correction, e.g., DivideMix \citep{li2020dividemix} conducts sample selection via a two-component gaussian mixture model and then applies the semi-supervised learning technique MixMatch with label correction. 

\noindent
\textbf{Learning from Long-Tailed Noisy Data.}
Methods designed for learning from long-tailed noisy data are under-explored. \citet{shu2019meta} and \citet{jiang2021delving} learned a weighting function from training data in a meta-learning manner, which can be applied to deal with long-tailed noisy data, yet these methods require extra unbiased meta-data. \citet{zhang2021learning} used a dictionary to store valuable training samples as a proxy of meta-data. 
In general, meta-learning methods attempt to learn a weighting function mapping training loss into sample weight and assign large weights to clean samples and tail samples. Nevertheless, in the learning process, clean samples usually have small losses and tail samples usually have large losses.
\citet{zhong2019unequal} proposed noise-robust loss functions and an unequal training strategy which treats head data and tail data differently for face recognition. They simply divided training data into head group and tail group without considering the class imbalance in each group.
\citet{cao2020heteroskedastic} proposed a heteroskedastic adaptive regularization, which regularizes different regions of input space differently. 
However, they assigned the same regularization strength to samples with the same observed label without considering label noise, which made the method sensitive to noisy data.
\citet{karthik2021learning} followed a two-stage training approach by using self-supervised learning to train an unbiased feature extractor and fine-tuning with an appropriate loss.
\citet{yi2022identifying} proposed an iterative framework H2E, which first trains a noise identifier invariant to the class and context distributional change and then learns a robust classifier. 
These methods use the observed class distribution in the training processes to handle class imbalance, while in real-world applications the class distribution is inaccessible under label noise.


\vspace{-0.1cm}
\section{Methodology}
\label{Methodology}
\vspace{-0.1cm}

Let $\mathcal{X}$ denote the instance space, for each $\bm x\in\mathcal{X}$, there exists a true label $y\in\mathcal{Y}=\{1,\dots, C\}$, where $C$ is the number of classes. Let $\tilde{D}=\{(\boldsymbol{x}_1, \tilde{y}_1), \dots , (\boldsymbol{x}_N, \tilde{y}_N)\}$ denote the training data,
where $\boldsymbol{x}_i \in \mathcal{X}$, $\tilde{y}_i \in \mathcal{Y}$ is the observed label of $\boldsymbol{x}_i$ that may be corrupted and $N$ is the number of training samples. In real-world applications, the data may follow a long-tailed distribution. Let $\tilde{D}_c=\{(\boldsymbol{x}_i, \tilde{y}_i) | (\boldsymbol{x}_i, \tilde{y}_i) \in \tilde{D}\land \tilde{y}_i=c\}$ and $n_c = |\tilde{D}_c|$. Without loss of generality, the classes are sorted by their cardinalities $n_c$ in the decreasing order, i.e., $n_1 > n_2 > \dots > n_C$, and the imbalance ratio $\rho=n_1/n_C$. 
The deep neural network $f(\cdot;\theta):\mathcal{X} \rightarrow \mathbb{R}^C$ is learned from the long-tailed and noisy data $\tilde{D}$, where $\theta$ represents the model parameter. Given $\bm x \in \mathcal{X}$, the network outputs $f(\bm x; \theta) = [f_1(\bm x;\theta), \dots ,f_C(\bm x;\theta)]^T$ and $f_i(\boldsymbol{x};\theta)$ represents the output logit of class $i$, we omit the model parameter $\theta$ and denote $f(\bm x;\theta)$ as $f(\bm x)$ for brevity. 
Let $p_{model}^i(\bm x) = \frac{\exp(f_i(\bm x))}{\sum_{j=1}^C \exp(f_j(\bm x))}$, the classifier induced by $f$ is $\phi_f(\bm x) = \mathop{\arg\max}_{i \in \{1,\dots,C\}} p_{model}^i(\bm x)$. For $(\bm x, \tilde{y})$, the loss is calculated as $\ell(f(\bm x), \tilde{y})$ with a loss fuction $\ell(\cdot,\cdot)$. The empirical loss of $f$ on $\tilde{D}$ is $\frac{1}{N}\sum_{i=1}^N \ell(f(\bm x_i), \tilde{y}_i)$. We focus on learning the optimal model parameter $\theta^*$ which minimizes the expected loss, i.e., $\theta^* = \mathop{\arg\min}_{\theta} \mathbb{E}_{(\bm x, y)} [\ell(f(\bm x;\theta), y)]$.




\floatname{algorithm}{Procedure}

\begin{algorithm}[t]
    \setcounter{algorithm}{0}
    \caption{Class-Aware Sample Selection (CASS)}
    \label{procedure1}

    \begin{algorithmic}[1]
    \Require Parameter of model $\theta$, training data $(\mathcal{X}, \mathcal{Y})$.
    
    \For{c = 1 to C}
        \State Draw $\tilde{D}_c = \{(\boldsymbol{x}_i, \tilde{y}_i)|(\boldsymbol{x}_i, \tilde{y}_i) \in (\mathcal{X}, \mathcal{Y})\land \tilde{y}_i = c\}$ from $\tilde{D} = (\mathcal{X}, \mathcal{Y})$
        \State $\mathcal{W}_c=\text{GMM}(\{\text{SoftmaxCE}(\boldsymbol{x}_i, \tilde{y}_i; \theta)\ |\ (\boldsymbol{x}_i,\tilde{y}_i) \in \tilde{D}_c\})$ \Comment{obtain samples' clean probabilities}
        \State $\mathcal{L}_c = \{(\boldsymbol{x}_i, \tilde{y}_i) | (\boldsymbol{x}_i, \tilde{y}_i, w_i) \in (\tilde{D}_c, \mathcal{W}_c) \land w_i > \frac{1}{2} \}$ 
        \State $\mathcal{U}_c = \{\boldsymbol{x}_i | (\boldsymbol{x}_i, \tilde{y}_i, w_i) \in (\tilde{D}_c, \mathcal{W}_c) \land w_i \leq \frac{1}{2} \}$ 
    \EndFor

    \Ensure $\mathcal{L}=\cup_{i=1}^{C} \mathcal{L}_i$, $\mathcal{U}=\cup_{i=1}^{C} \mathcal{U}_i$.

    \end{algorithmic}
\end{algorithm}



The training data in real-world applications usually contain noisy labels, while deep neural networks are generally learned by minimizing the empirical risk on the training data. \citet{zhang2021understanding} demonstrated that
noisy labels can be easily fitted by deep neural networks, which harms the generalization of the neural networks. In order to alleviate the effect of noisy labels, we adopt the popular strategy which first warms up the model and then selects clean samples with small-loss criterion \citep{han2018co,li2020dividemix}. Since the training data are long-tailed in real-world applications, we introduce a regularization term $L_{reg}$ in the warm up process to prevent the model from being influenced by the long-tailed distribution:
\begin{equation}
    L_{reg} = \sum_{i=1}^C \frac{n_C}{n_i} \pi_i \log \left(\pi_i \bigg/ \frac{1}{C} \sum_{j=1}^C \sum_{(\bm x, \tilde{y}) \in \tilde{D}_j} \frac{1}{n_j} p_{model}^i(\bm x)\right),
    \nonumber
\end{equation}
where $\pi_i = \frac{1}{C}$. $L_{reg}$ forces the model's average output on all classes to be the uniform distribution. Considering that head classes have more samples and the calculation of the model's average output may be biased, we conduct the calculation in a class-balanced manner by assigning a weight $\frac{1}{n_j}$ to samples from class $j$. We pay more attention to tail classes by assigning a regularization strength $\frac{n_C}{n_i}$ to class $i$, which guarantees a decent performance on tail classes.
During warming up, hyper-parameter $\lambda_{warm}$ controls the strength of $L_{reg}$:
\begin{equation}
    \label{warmup}
    L = L_{CE} + \lambda_{warm}L_{reg}.
\end{equation}

After warming up, small-loss criterion can be used to select small-loss samples as clean ones following \citet{arpit2017a}.
Recently, \citet{gui2021towards} revealed that the losses of samples with different labels may not be comparable and proposed to select samples class by class accordingly. They set each class a threshold based on the noise rate. Unfortunately, each class's noise rate is unknown in practice. 
Instead, we adopt a two-component ($g_1$ and $g_0$) Gaussian Mixture Model (GMM) for each class in the sample selection process to fit the loss as that in \citet{arazo2019unsupervised} and \citet{li2020dividemix}, where $g_1$ represents the clean distribution and $g_0$ represents the noisy distribution. For class $i$, let $L(\tilde{D}_i)$ denote the samples' cross-entropy loss, and it reflects how well the model fits the training samples:
\begin{equation}
    L(\tilde{D}_i) = \left\{-\log\left(\frac{\exp(f_i(\bm x_j))}{\sum_{k=1}^{C} \exp(f_k(\bm x_j))}\right)\bigg|(\bm x_j, \tilde{y}_j) \in \tilde{D}_i\right\}_{j=1}^{n_i}.
    \nonumber
\end{equation}
A two-component GMM can be fitted with respect to $L(\tilde{D}_i)$, and
the clean probability of a sample $(\bm x_j, \tilde{y}_j)$ is given by its posterior probability $P(g_1|L(\tilde{D}_i), (\bm x_j, \tilde{y}_j))$. The sample with clean probability larger than $1/2$ is selected as clean; otherwise, the sample is regarded as noisy.
Here, we do not need to know the proportion of clean samples which depends on the inaccessible noise rate. 
In this way, the clean labeled set $\mathcal{L}$ can be selected from the training data $\tilde{D}$, and an unlabeled set $\mathcal{U} = \{\bm x_i | (\bm x_i, \tilde{y}_i) \in \tilde{D} \backslash \mathcal{L}\}$ is constructed. This class-aware sample selection is described in Procedure \ref{procedure1}.
We further plot the loss distributions of different classes in order to demonstrate that the distributions follow the two-component GMM (see Appendix \ref{appendix:visualize} for details).

With clean labeled samples $\mathcal{L}$ and unlabeled samples $\mathcal{U}$, the widely-used semi-supervised learning method MixMatch \citep{berthelot2019mixmatch} can be adopted to learn the model. MixMatch uses the learned model to generate pseudo labels for $\mathcal{U}$, applies MixUp to transform $\mathcal{L}$ and $\mathcal{U}$ into $\mathcal{L}'$ and $\mathcal{U}'$ with soft labels $\bm q \in [0, 1]^C$ and then utilizes
the cross-entropy loss and the mean squared error on $\mathcal{L}'$ and $\mathcal{U}'$ respectively:
\vspace{-0.3cm}

\begin{align}
    L_{\mathcal{L}} = \frac{1}{|\mathcal{L}'|} \sum_{(\bm x, \bm q) \in \mathcal{L}'} L(\bm x, \bm q)
    &= -\frac{1}{|\mathcal{L}'|} \sum_{(\bm x, \bm q) \in \mathcal{L}'} \sum_{i=1}^C q_i \log \frac{\exp(f_i(\bm x))}{\sum_{j=1}^C \exp(f_j(\bm x))} \label{ce} \\
    &= \frac{1}{|\mathcal{L}'|} \sum_{(\bm x, \bm q) \in \mathcal{L}'} \sum_{i=1}^C q_i \log \left[1 + \sum_{j \neq i} \frac{\exp(f_j(\boldsymbol{x}))}{\exp(f_i(\boldsymbol{x}))}\right], \nonumber
\end{align}
\begin{equation}
    L_{\mathcal{U}} = \frac{1}{|\mathcal{U}'|} \sum_{(\bm x, \bm q) \in \mathcal{U}'} \|\bm q - p_{model}(\bm x)\|_2^2.
\end{equation}

\floatname{algorithm}{Algorithm}

\begin{algorithm*}[ht]
    
    \renewcommand{\thealgorithm}{1}
    \caption{Learning with class-aware Sample Selection and Balanced Loss (SSBL).}
    \label{main1}

    \begin{algorithmic}[1]
        \Require Deep neural network $f(\cdot; \theta)$, training data $(\mathcal{X}, \mathcal{Y})$, total training epochs $T$, model bias estimation epochs $E$, unsupervised loss weight $\lambda_u$, regularization term weight $\lambda_{reg}$, hyper-paramter of EMA $\sigma$.
        
        \State $\theta$ = WarmUp$(\mathcal{X}, \mathcal{Y}, \theta)$ \Comment{training with Eq. (\ref{warmup})}
        \State Fill $\bar{M} \in \mathbb{R}^{C \times C}$ with zeros
        \While{$e < E$}
            \State Fill $M^e \in \mathbb{R}^{C \times C}$ with zeros
            \State ($\mathcal{L}, \mathcal{U}$) = CASS($\mathcal{X}, \mathcal{Y}, \theta$) \Comment{apply Procedure \ref{procedure1}}
            \For{iter = 1 to num\_iters}
                \State Draw a mini-batch $\mathcal{L}_B = \{(\bm x_b, \tilde{y}_b)|b \in \{1,...,B\}\}$ from $\mathcal{L}$
                \State Draw a mini-batch $\mathcal{U}_B = \{\bm u_b|b \in \{1,...,B\}\}$ from $\mathcal{U}$
                
                \For{c = 1 to C}
                    \State Draw $\mathcal{L}_{B, c} = \{(\bm x, \tilde{y})|(\bm x, \tilde{y}) \in \mathcal{L}_{B}\land \tilde{y} = c\}$ from $\mathcal{L}_{B}$
                    \State $M^e_c = M^e_c + \sum_{(\bm x, \tilde{y}) \in \mathcal{L}_{B, c}}$ Softmax$(f(\bm x; \theta))$ \Comment{update the c-th row of $M^e$}
                \EndFor
                
                \State ($\mathcal{L}', \mathcal{U}'$) = $\text{MixMatch}$$(\mathcal{L}_B, \mathcal{U}_B)$ \Comment{apply $\text{MixMatch}$ with augmented $\mathcal{L}_B$ and $\mathcal{U}_B$}
                \State $L_\mathcal{L}, L_\mathcal{U}$ = SoftmaxCE$(\mathcal{L}'; \theta)$, SoftmaxMSE$(\mathcal{U}'; \theta)$
                \State $L = L_\mathcal{L} + \lambda_u L_\mathcal{U} + \lambda_{reg} L_{reg}$
                \State $\theta = SGD(L, \theta)$ \Comment{update the parameters of $f(\cdot;\theta)$ with Stochastic Gradient Descent}
            \EndFor
            \State Normalize $M^e$
            \State $\bar{M} = \sigma \bar{M} + (1-\sigma) M^e$ \Comment{update the model bias matrix $\bar{M}$}
        \EndWhile
        
        \State $R_{ij} = \frac{\bar{M}_{ij}}{\bar{M}_{ji}}, R \in \mathbb{R}^{C \times C}$ \Comment{compute $R$}
        
        \While{$e < T$}
            \State ($\mathcal{L}, \mathcal{U}$) = CASS($\mathcal{X}, \mathcal{Y}, \theta$) \Comment{apply Procedure \ref{procedure1}}
            \For{iter = 1 to num\_iters}
                \State Draw a mini-batch $\mathcal{L}_B = \{(\bm x_b, \tilde{y}_b)|b \in \{1,...,B\}\}$ from $\mathcal{L}$
                \State Draw a mini-batch $\mathcal{U}_B = \{\bm u_b|b \in \{1,...,B\}\}$ from $\mathcal{U}$
                
                \State ($\mathcal{L}', \mathcal{U}'$) = $\text{MixMatch}$$(\mathcal{L}_B, \mathcal{U}_B)$ \Comment{apply $\text{MixMatch}$ with augmented $\mathcal{L}_B$ and $\mathcal{U}_B$}
                \State $L'_\mathcal{L}, L_\mathcal{U} = L_{\bm \alpha}(\mathcal{L}'; \theta, R)$, SoftmaxMSE$(\mathcal{U}'; \theta)$
                \State $L = L'_\mathcal{L} + \lambda_u L_\mathcal{U} + \lambda_{reg} L_{reg}$
                \State $\theta = SGD(L, \theta)$ \Comment{update the parameters of $f(\cdot;\theta)$ with Stochastic Gradient Descent}
            \EndFor
        \EndWhile

    \end{algorithmic}
\end{algorithm*}

\noindent In Eq. (\ref{ce}) of MixMatch, training with long-tailed data leads to model bias, and the model with bias prefers to predict head classes and has a poor performance on tail classes. 
This motivates us to develop a loss to correct the model bias. 
That is, we introduce $\alpha_{ij}$ into $L(\bm x, \bm q)$ in Eq. (\ref{ce}):
\begin{equation}
\begin{split}
  L_{\bm \alpha}(\bm x, \bm q) &= -\sum_{i=1}^C q_i \log \frac{ \exp(f_i(\boldsymbol{x}))}{\sum_{j=1}^C \alpha_{ij} \exp(f_j(\boldsymbol{x}))}
  = \sum_{i=1}^C q_i \log \left[\alpha_{ii} + \sum_{j \neq i} \alpha_{ij} \frac{\exp(f_j(\boldsymbol{x}))}{\exp(f_i(\boldsymbol{x}))}\right].
\end{split}
\end{equation}
Specifically, for a pair of classes $(i, j)$, if the learned model prefers class $j$, then a weight $\alpha_{ij}>1$ is assigned to $\frac{\exp(f_j(\boldsymbol{x}))}{\exp(f_i(\boldsymbol{x}))}$ for samples from class $i$ to suppress class $j$, and a weight $\alpha_{ji}<1$ is assigned to $\frac{\exp(f_i(\boldsymbol{x}))}{\exp(f_j(\boldsymbol{x}))}$ for samples from class $j$ to relax class $i$. 
Many methods use label frequency as the estimation of model bias in long-tailed learning \citep{cui2019cbloss,menon2020long}. However, the model bias on different classes is not directly related to label frequency (see Appendix \ref{appendix:frequency}), and the true label frequency is also unknown under label noise.
To characterize the model bias, we calculate the matrix $M \in \mathbb{R}^{C \times C}$ before each training epoch, where the entry $M_{ij}$ represents the probability that the learned model predicts a sample from class $i$ to class $j$. For epoch $t$,
\begin{equation}
  M^t_{ij} = \frac{1}{|\mathcal{L}_i|} \sum_{(\boldsymbol{x}, \tilde{y}) \in \mathcal{L}_i} p_{model}^{j, t}(\bm x),
\end{equation}
where $\mathcal{L}_i = \{(\bm x, \tilde{y})|(\bm x, \tilde{y}) \in \mathcal{L}\land \tilde{y} = i\}$. Inspired by \citet{samuli2017temporal}, we further temporally average the calculated matrices for a stable estimation of model bias with Exponentially Moving Average (EMA) and obtain the averaged model bias matrix $\bar{M}$, i.e.,
$\bar{M}^t = \sigma \bar{M}^{t-1} + (1-\sigma) M^t$,
where $\sigma$ controls the contribution of the estimated matrix from each epoch.
If $\bar{M}_{ij}$ is larger than $\bar{M}_{ji}$, then the model has a bias to class $j$. 
Let $R_{ij} = \bar{M}_{ij} / \bar{M}_{ji}$, $g(\cdot)=  \gamma_{sup} \cdot \mathbb{I}(R_{ij}>1) + \gamma_{rel} \cdot \mathbb{I}(R_{ij} \leq 1)$ and $\alpha_{ij} = R_{ij}^{g(R_{ij})}$, where $g(\cdot)$ is a function of $R_{ij}$ in which $\gamma_{sup}$ and $\gamma_{rel}$ are hyper-parameters for controlling the power of suppressing and relaxing respectively.
Formally, we have
\begin{equation}
\nonumber
    \begin{split}
    L_{\mathcal{L}}' = \frac{1}{|\mathcal{L}'|} \sum_{(\bm x, \bm q) \in \mathcal{L}'} L_{\bm \alpha}(\bm x, \bm q) 
    &= -\frac{1}{|\mathcal{L}'|} \sum_{(\bm x, \bm q) \in \mathcal{L}'} \sum_{i=1}^C q_i \log \frac{\exp(f_i(\bm x))}{\sum_{j=1}^C R_{ij}^{g(R_{ij})} \exp(f_j(\bm x))} \\
    &= \frac{1}{|\mathcal{L}'|} \sum_{(\bm x, \bm q) \in \mathcal{L}'} \sum_{i=1}^C q_i \log \left[1+\sum_{j \neq i} R_{ij}^{g(R_{ij})} \frac{\exp(f_j(\bm x))}{\exp(f_i(\bm x))}\right].
    \end{split}
\end{equation}

Since the training data are in long-tailed distribution and there are only a few samples in tail classes, we augment clean labeled samples in $\mathcal{L}$ and re-exploit $L_{reg}$ as a regularization on $\mathcal{L}'$ with its hard labels.
Thus, we use the following balanced loss in the training process:
\begin{equation}
    L = L'_{\mathcal{L}} + \lambda_u L_{\mathcal{U}} + \lambda_{reg} L_{reg},
\end{equation}
where $\lambda_u$ controls the strength of the unsupervised loss and $\lambda_{reg}$ controls the strength of the regularization term.
The whole training process of our method SSBL is described in Algorithm \ref{main1}.

\begin{table*}[ht]
\Huge

    \centering
    \resizebox{400pt}{!}{%
    \begin{tabular}{l|c|c c c|c c c|c c c|c c c}
    \toprule
    \multicolumn{2}{c|}{Dataset} & \multicolumn{6}{c|}{CIFAR-10} & \multicolumn{6}{c}{CIFAR-100} \\
      \hline
    \multicolumn{2}{c|}{Noise Rate} & \multicolumn{3}{c|}{0.2} & \multicolumn{3}{c|}{0.5} & \multicolumn{3}{c|}{0.2} & \multicolumn{3}{c}{0.5}\\
      \hline
    \multicolumn{2}{c|}{Imbalance Ratio} & 10    & 50    & 100   & 10    & 50    & 100     & 10    & 50    & 100   & 10    & 50    & 100\\
      \hline
    \multirow{2}{*}{{ERM}} &  {Best} &  {76.90} &  {65.35} &  {60.82} &  {65.75} &  {48.76} &  {39.70} & {45.83} & {35.05} & {29.96} & {28.96} & {19.88} & {16.80}\\
           &  {Last} &  {73.02} &  {61.35} &  {54.48} &  {45.85} &  {33.05} &  {28.79} & {45.64} & {34.93} & {29.88} & {24.33} & {17.77} & {14.47} \\
      \hline
    \multirow{2}{*}{Co-teaching \citep{han2018co}} &  {Best} &  {78.50} &  {45.86} &  {39.59} &  {34.60} &  {23.58} &  {17.45} & {43.81} & {30.58} & {28.08} & {14.58} & {11.62} & {9.69} \\
           &  {Last} &  {77.51} &  {44.91} &  {38.07} &  {31.71} &  {22.57} &  {14.88} & {43.69} & {30.22} & {28.08} & {14.49} & {10.54} & {9.35} \\
      \hline
    \multirow{2}{*}{HAR-DRW \citep{cao2020heteroskedastic}} & Best  & 73.83 & 61.30  & 60.92 & 59.10  & 43.69 & 36.31 & 37.31 & 29.29 & 26.00    & 23.22 & 17.04 & 12.45 \\
           & Last  & 71.16 & 60.27 & 56.94 & 40.10  & 34.19 & 35.91 & 36.92 & 29.02 & 25.69 & 19.01 & 14.29 & {11.11} \\
      \hline
      
    \multirow{2}{*}{MW-Net \citep{shu2019meta}} &  
    {Best} & 83.78	& 67.97	& 50.96	& 71.81	& 38.45	& 27.15	& 51.09	& 37.88	& 33.41	& 31.83	& 18.68	& 14.34 \\
    
    &  {Last} & 73.25 & 62.63 & 48.01 & 50.02 & 36.14 & 20.90 & 50.24 & 37.75	& 33.24 & 27.72	& 13.61	& 13.20 \\
      \hline

    \multirow{2}{*}{H2E \citep{yi2022identifying}} &  
    {Best} & 79.40 & 59.49 & 52.80 & 62.03 & 36.29 & 31.79 & 48.66 & 34.86 & 29.26 & 33.38 & 22.92 & 19.15 \\
    
    &  {Last} & 78.76 & 52.10 & 49.95 & 43.66 & 33.57 & 31.10 & 48.48 & 34.65 & 28.95 & 33.17 & 19.73 & 14.69 \\
      \hline

    \multirow{2}{*}{SSBL} & Best & \textbf{91.60} & \textbf{86.30} & \textbf{79.84} & \textbf{88.51} & \textbf{77.72} & \textbf{72.36} & \textbf{62.78} & \textbf{51.25} & \textbf{45.68} & \textbf{55.95} & \textbf{42.19} & \textbf{36.87}\\
    & Last &\textbf{91.49} & \textbf{85.83} & \textbf{78.39} & \textbf{88.17} & \textbf{75.76} & \textbf{68.44} & \textbf{62.45} & \textbf{50.70} & \textbf{43.89} & \textbf{55.65} & \textbf{40.82} & \textbf{36.49} \\
    
     \hline
     \hline
     
    \multirow{2}{*}{ELR+ \citep{liu2020early}} &  {Best} &  {88.34} &  {76.31} &  {68.39} &  {64.61} &  {16.40} &  {15.65} & {53.11} & {39.17} & {34.35} & {35.97} & {25.27} & {20.49} \\
           &  {Last} &  {87.68} &  {74.62} &  {66.60} &  {60.30} &  {10.03} &  {10.03} & {48.95} & {35.20} & {31.36} & {22.73} & {18.78} & {16.36} \\
      \hline
     
    \multirow{2}{*}{DivideMix \citep{li2020dividemix}} &  {Best} & 91.03 & 83.07 & 70.08 & 85.97 & 69.73 & 52.06 & 63.08 & 49.76 & 43.71 & 55.98 & 41.79 & 35.03 \\
           &  {Last} & 90.71 & 82.16 & 69.82 & 85.79 & 68.19 & 51.72 & 62.54 & 49.45 & 42.84 & 55.56 & 41.25 & 34.51 \\
      \hline

    \multirow{2}{*}{DivideMix-LA} &  {Best} & 91.68 & 85.16 & 80.06 & 85.84 & 56.28 & 49.48 & {64.76} & 53.31 & 47.52 & 57.11 & 40.72 & 35.18\\
 & {Last} & 91.65 & 84.15 & 78.40 & 85.47 & 54.67 & 46.66 & {64.33} & 52.83 & 46.21 & 56.57 & 40.21 & 34.73\\
      \hline
      
    \multirow{2}{*}{DivideMix-DRW} &  {Best} &90.44	&84.64	& 80.50	&85.40	&73.90	&47.31 & 63.65 	&50.28 	&45.23 	&56.77 &42.24 	&36.17 \\
           &  {Last} &89.33	&82.99	&80.40	&84.88	&72.92	&45.90 & 63.24 	&50.04 	&44.95 	&56.58 	&41.61 	&35.74
 \\
      \hline

    \multirow{2}{*}{$\text{SSBL}_2$} &  {Best} & \textbf{92.47} & \textbf{87.14} & \textbf{81.98} & \textbf{89.41} & \textbf{78.65} & \textbf{72.96} & \textbf{65.09} & \textbf{53.52} & \textbf{47.87} & \textbf{57.95} & \textbf{44.37} & \textbf{39.49} \\
           &  {Last} &\textbf{92.25}	&\textbf{86.87}	& \textbf{81.29}	&\textbf{89.09}	&\textbf{76.44}	&\textbf{69.11} & \textbf{64.60} & \textbf{53.35} & \textbf{47.64} & \textbf{57.80} & \textbf{43.47} & \textbf{38.64}
 \\
    \bottomrule
    \end{tabular}%
    }
   \caption{Comparison with baselines in test accuracy (\%) on long-tailed versions of CIFAR-10 and CIFAR-100 with symmetric noise. $\text{SSBL}$ represents our performance of single model, while $\text{SSBL}_2$ represents our ensemble performance of two models.}
   \label{cifar-sym}%
\end{table*}%



\section{Experiment}

\subsection{Setup}
\label{setup}

\textbf{Datasets.}
We validate our method on seven benchmark datasets, namely CIFAR-10, CIFAR-100 \citep{krizhevsky2009learning}, mini-ImageNet-Red \citep{jiang2020beyond}, Clothing1M \citep{xiao2015learning}, Food-101N \citep{lee2018cleannet}, Animal-10N \citep{song2019selfie} and WebVision \citep{Li2017a}. On CIFAR, we consider two kinds of label noise, i.e., symmetric noise and asymmetric noise. For symmetric noise, we first construct long-tailed versions of CIFAR with different imbalance ratios $\rho$ following \citet{cui2019cbloss}. In specific, we reduce the sample size per class according to an exponential fuction $n_i = O_i \frac{1}{\rho}^{\frac{i-1}{C-1}}$, where $O_i$ is the original sample size of class $i$, $\rho = \frac{n_1}{n_C}$ and $i \in \{1,\dots,C\}$.
Then, we generate symmetric noise in long-tailed CIFAR according to the following noise transition matrix $N$:
\begin{equation}
    N_{ij}(\bm x, y) = \mathcal{P}(\tilde{y}=j|y=i, \bm x) = \left\{
    \begin{aligned}
        &1-r & i&=j \\
        &r \frac{n_j}{\sum_{k\neq i}n_k} & i&\neq j,
    \end{aligned}
    \right.
    \nonumber
\end{equation}
where $y$ denotes the ground truth of $\bm x$, $\tilde{y}$ denotes the corrupted label of $\bm x$ and $r \in [0, 1]$ denotes the noise rate. We set imbalance ratio to be $\rho \in \{10, 50, 100\}$ and set noise rate to be $r \in \{0.2, 0.5\}$.
In the experiments of learning from long-tailed noisy data, the noise rate is usually lower than 0.5, as that in \citet{zhang2021learning} and \citet{cao2020heteroskedastic}.
For asymmetric noise, we construct step-imbalanced versions of CIFAR-10 with asymmetric noise following \citet{cao2020heteroskedastic}. In specific, we corrupt semantically-similar classes by exchanging 40\% of the labels between `cat' and `dog', and by exchanging 40\% of the labels between `truck' and `automobile'. Then, we remove samples from the corrupted classes. Here, the imbalance ratio $\rho$ is the sample size ratio between the frequent (and clean) classes and the rare (and noisy) classes. 
Mini-ImageNet-Red, Clothing1M, Food-101N, Animal-10N and WebVision are datasets with real-world noise, so we directly create their long-tailed versions following \citet{cui2019cbloss}.
For mini-ImageNet-Red, Clothing1M, Food-101N and Animal-10N, we set imbalance ratio to be $\rho \in \{20, 50, 100\}$.
For Clothing1M, we randomly select a small balanced subset of Clothing1M following \citet{yi2019probabilistic} and construct its long-tailed versions.
For WebVision, we train and test on the subset mini WebVision which contains the first 50 classes of WebVision following \citet{li2020dividemix}. 
The original imbalance ratio of mini WebVision is approximately 7, we further create long-tailed versions of mini WebVision with $\rho \in \{50, 100\}$ following \citet{cui2019cbloss}.



\begin{table}[t]
\centering%
\begin{minipage}[t]{162pt}

    \large
    \centering
    \resizebox{162pt}{!}{%
    \begin{tabular}{l|c c c}
    \toprule
    {Imbalance Ratio} & {10}  &  {50}   & {100} \\
    \hline
    {ERM} &  {77.45} & {71.01} & {69.00} \\
     \hline
    {Co-teaching \citep{han2018co}} &  {77.06} & {69.49} & {67.34}  \\
    \hline
    {HAR-DRW \citep{cao2020heteroskedastic}} & {82.69} &  {72.17} & {69.32}  \\
    \hline
    {MW-Net \citep{shu2019meta}} & {81.42} & {72.82} & {68.92}  \\
    \hline
    {H2E \citep{yi2022identifying}} & 76.80 &  68.89 &  68.08 \\
    \hline
    {SSBL} &  \textbf{87.88}	& \textbf{72.95}	& \textbf{70.16} \\
    \hline
    \hline
    {ELR+ \citep{liu2020early}} & {74.09}	& {68.33}	& {67.04} \\
    \hline
    {DivideMix \citep{li2020dividemix}} & {88.07} & {69.19} & {67.46} \\
    \hline
    {DivideMix-LA} & 83.88 & 72.11 & 66.41 \\
    \hline
    {DivideMix-DRW} & {81.56} & {71.02} & {69.80} \\
    \hline
    {$\text{SSBL}_2$} & \textbf{88.83} & \textbf{73.96} & \textbf{71.43} \\
    \bottomrule
    \end{tabular}%
    }
    \vspace{0.2cm}
    \caption{Comparison with baselines in test accuracy (\%) on step-imbalanced versions of CIFAR-10 with asymmetric noise. $\text{SSBL}$ represents our performance of single model, while $\text{SSBL}_2$ represents our ensemble performance of two models.}
    \label{cifar-asym-only-all}%
\end{minipage}%
\hspace{14pt}%
\begin{minipage}[t]{215pt}
    \large
    \centering
    \resizebox{218pt}{!}{
    \begin{tabular}{l|c c c|c c c}
    \toprule
    {Noise Rate} & \multicolumn{3}{c|}{0.2} & \multicolumn{3}{c}{0.5} \\
    \hline
    {Imbalance Ratio} & {20} & {50} & {100} & {20} & {50} & {100} \\
    \hline
    {Co-teaching \citep{han2018co}} & {22.72} & {20.09} & {17.05} & {17.97} & {14.74} & {13.86} \\
    \hline
    {HAR-DRW \citep{cao2020heteroskedastic}} & {29.78} & {25.94} & {23.30} & {24.10} & {21.34} & {19.14} \\
    \hline
    {MW-Net \citep{shu2019meta}} & {33.56} & {27.74} & {24.46} & {26.72} & {23.10} & {20.82} \\
    \hline
    {H2E \citep{yi2022identifying}} & {26.26} & {24.56} & {20.82} & {25.76} & {20.50} & {17.80} \\
    \hline
    {SSBL} & {36.02} & {31.38} & {28.20} & {30.04} & {26.36} & {23.98} \\
    \hline
    {SSBL+RandAug} & \textbf{37.52} & \textbf{32.54} & \textbf{29.54} & \textbf{31.14} & \textbf{27.94} & \textbf{25.34} \\
    \hline
    \hline
    {ELR+ \citep{liu2020early}} & {31.86} & {26.33} & {24.39} & {26.23} & {23.30} & {20.61} \\
    \hline
    {DivideMix \citep{li2020dividemix}} & {35.96} & {29.02} & {27.36} & {30.10} & {26.42} & {24.20} \\
    \hline
    {DivideMix-LA} & {37.36} & {32.12} & {29.66} & {32.34} & {28.12} & {24.78} \\
    \hline
    {DivideMix-DRW} & {36.56} & {30.14} & {28.48} & {30.64} & {26.62} & {24.34} \\
    \hline
    {$\text{SSBL}_2$} & \textbf{37.92} & \textbf{32.90} & \textbf{30.74} & \textbf{32.84} & \textbf{29.06} & \textbf{26.26} \\
    \bottomrule
    \end{tabular}
    }
    
    \vspace{0.25cm}
    \caption{Comparison with baselines in test accuracy (\%) on long-tailed versions of mini-ImageNet-Red with real-world noise. $\text{SSBL}$ represents our performance of single model, while $\text{SSBL}_2$ represents our ensemble performance of two models.}
    \label{mini-imagenet-red}

\end{minipage}
\end{table}

\begin{table*}[t]
\tiny

    \centering
    \resizebox{400pt}{!}{
    \begin{tabular}{l|c c c|c c c|c c c}
    \toprule
    {Dataset} & \multicolumn{3}{c|}{Clothing1M} & \multicolumn{3}{c|}{Food-101N} & \multicolumn{3}{c}{Animal-10N} \\
    \hline
    {Imbalance Ratio} & {20} & {50} & {100} & {20} & {50} & {100} & {20} & {50} & {100} \\
    \hline
    {Co-teaching \citep{han2018co}} & {51.97} & {43.23} & {36.48} & {45.67} & {38.90} & {35.13} & {55.06} & {34.51} & {32.63} \\
    \hline
    {HAR-DRW \citep{cao2020heteroskedastic}} & {62.56} & {54.08} & {51.09} & {54.12} & {48.25} & {41.81} & {71.96} & {64.22} & {57.66} \\
    \hline
    {MW-Net \citep{shu2019meta}} & {60.04} & {57.15} & {51.98} & {56.73} & {47.44} & {41.89} & {74.72} & {64.26} & {53.86} \\
    \hline
    {H2E \citep{yi2022identifying}} & {70.41} & {67.81} & {61.58} & {70.35} & {63.69} & {58.66} & {77.04} & {74.94} & {66.58} \\
    \hline
    {SSBL} & \textbf{71.74} & {69.37} & {66.80} & {72.05} & {66.78} & {61.62} & {80.20} & {75.06} & {69.10} \\
    \hline
    {SSBL+RandAug} & {71.66} & \textbf{71.04} & \textbf{67.65} & \textbf{74.06} & \textbf{68.56} & \textbf{64.64} & \textbf{81.58} & \textbf{78.82} & \textbf{71.84} \\
    \hline
    \hline
    {ELR+ \citep{liu2020early}} & {67.46} & {61.24} & {54.83} & {58.56} & {48.98} & {44.05} & {63.37} & {49.28} & {47.53} \\
    \hline
    {DivideMix \citep{li2020dividemix}} & {71.38} & {69.27} & {66.63} & {64.02} & {54.65} & {52.57} & {77.64} & {69.74} & {64.72} \\
    \hline
    {DivideMix-LA} & {71.20} & {70.09} & {67.20} & {71.48} & {63.73} & {59.01} & {78.62} & {74.90} & {66.08} \\
    \hline
    {DivideMix-DRW} & {71.56} & {69.83} & {66.85} & {64.21} & {53.98} & {53.65} & {79.32} & {74.14} & {66.14} \\
    \hline
    {$\text{SSBL}_2$} & \textbf{72.41} & \textbf{70.76} & \textbf{67.91} & \textbf{74.39} & \textbf{69.06} & \textbf{63.67} & \textbf{81.20} & \textbf{75.84} & \textbf{69.36} \\
    \bottomrule
    \end{tabular}
    }
    
    \caption{Comparison with baselines in test accuracy (\%) on long-tailed versions of Clothing1M, Food-101N and Animal-10N with real-world noise. $\text{SSBL}$ represents our performance of single model, while $\text{SSBL}_2$ represents our ensemble performance of two models. Results of H2E on Food-101N and Animal-10N are borrowed from its original paper.}
    \label{clothing1m_food101n_animal10n}
\end{table*}

\vspace{0.2cm}
\noindent
\textbf{Baselines.}
We compare our method with state-of-the-art label noise learning methods and long-tailed label noise learning methods, including Empirical Risk Minimization (ERM) which trains the model with the cross-entropy loss, Co-teaching \citep{han2018co} which trains two networks simultaneously and updates one network on the data selected by the other with small-loss criterion, ELR+ \citep{liu2020early} which capitalizes on early learning via regularization, DivideMix \citep{li2020dividemix} which conducts sample selection via a two-component gaussian mixture model and then applies the semi-supervised learning technique MixMatch with label correction, HAR-DRW \citep{cao2020heteroskedastic} which regularizes different regions of input space differently to handle heteroskedasticity and imbalance issues simultaneously, MW-Net \citep{shu2019meta} which learns a weighting function in a meta-learning manner to assign each sample a weight, 
H2E \citep{yi2022identifying} which first trains a noise identifier and then learns a robust classifier, 
DivideMix-LA which combines DivideMix and Logit Adjustment (LA) \citep{menon2020long}, 
and DivideMix-DRW which combines DivideMix and Deferred Re-Weighting (DRW).





\noindent
\textbf{Implementation Details.}
On CIFAR-10, CIFAR-100, mini-ImageNet-Red and Animal-10N, we use an 18-layer PreAct ResNet and train for 200 epochs.
On Clothing1M and Food-101N, we use a ResNet-50 and train for 200 epochs from scratch.
On WebVision, we use an Inception-ResNet v2 and train for 100 epochs following \citet{li2020dividemix}.
Since the baseline H2E trains for 200 epochs in its original paper, we also conduct experiments on WebVision with H2E for 200 epochs.
On all datasets, $\gamma_{sup}$ is set as 3 and $\gamma_{rel}$ is set as 1 in $L'_{\mathcal{L}}$.
Please refer to Appendix \ref{appendix:setup} for more details.
The baselines ELR+, DivideMix, DivideMix-LA and DivideMix-DRW are based on two models.
We also run our method SSBL twice with random initialization and use the ensemble of two runs for a fair comparison with them. We refer to our method with two models as $\text{SSBL}_2$ and also report its performance in Tables \ref{cifar-sym}, \ref{cifar-asym-only-all}, \ref{mini-imagenet-red}, \ref{clothing1m_food101n_animal10n} and \ref{webvision-all}.
For a fair comparison with H2E, which uses strong augmentation technique RandAugment \citep{cubuk2020randaugment} in training, we also report the performance of SSBL with RandAugment (SSBL+RandAug).

\subsection{Evaluation on Benchmarks}



Table \ref{cifar-sym} summarizes the results on long-tailed versions of CIFAR-10 and CIFAR-100 with symmetric noise, and Table \ref{cifar-asym-only-all} summarizes the results on step-imbalanced CIFAR-10 with asymmetric noise.
From these tables, it can be found that SSBL with single model outperforms ERM, Co-teaching, HAR-DRW, MW-Net and H2E under all settings, and $\text{SSBL}_2$ with two models outperforms ELR+, DivideMix, DivideMix-LA and DivideMix-DRW under all settings.


Tables \ref{mini-imagenet-red} and \ref{clothing1m_food101n_animal10n} summarize the results on long-tailed mini-ImageNet-Red, Clothing1M, Food-101N and Animal-10N with real-world noise. From these tables, it can be found that SSBL with single model outperforms Co-teaching, HAR-DRW, MW-Net and H2E under all settings, and its performance can be further boosted with RandAugment. $\text{SSBL}_2$ with two models also outperforms ELR+, DivideMix, DivideMix-LA and DivideMix-DRW under all settings.

Table \ref{webvision-all} summarizes the results on original and long-tailed versions of mini WebVision with real-world noise, where we report Top-1 and Top-5 test accuracy on both WebVision and ImageNet following \citet{li2020dividemix}. From Table \ref{webvision-all}, it can be found that SSBL with single model outperforms Co-teaching, HAR-DRW, MW-Net and H2E in Top-1 test accuracy under all settings, and has the comparable or better performance in Top-5 test accuracy.
The performance of SSBL with single model can be further boosted with RandAugment.
$\text{SSBL}_2$ with two models also outperforms ELR+, DivideMix, DivideMix-LA and DivideMix-DRW in Top-1 test accuracy under all settings, and has the comparable or better performance in Top-5 test accuracy.

\begin{table*}[t]
\Huge

    \centering
    \resizebox{400pt}{!}{
    \begin{tabular}{l|c|c|c|c|c|c}
    \toprule
    {Imbalance Ratio} & \multicolumn{2}{c|}{Original} & \multicolumn{2}{c|}{50} & \multicolumn{2}{c}{100} \\
    \hline
    {Dataset} & {WebVision} & {ImageNet} & {WebVision} & {ImageNet} & {WebVision} & {ImageNet} \\
    \hline
    {Co-teaching \citep{han2018co}} & {63.60 (85.20)} & {61.50 (84.70)} & {43.11 (56.13)} & {40.95 (55.69)} & {40.42 (52.72)} & {37.78 (51.16)}\\
    \hline
    {HAR-DRW \citep{cao2020heteroskedastic}} & {72.96 (90.12)} & {67.12 (89.36)} & {58.16 (82.60)} & {54.20 (80.80)} & {50.24 (78.16)} & {47.44 (75.96)}\\
    \hline
    {MW-Net \citep{shu2019meta}} & {71.76 (90.40)} & {67.36 (89.08)} & {60.12 (84.08)} & {55.20 (82.28)} & {49.96 (80.04)} & {46.84 (77.80)}\\
    \hline
    {H2E \citep{yi2022identifying}} & {72.32 (90.84)} & {69.24 (\textbf{91.28})} & {57.88 (84.76)} & {56.64 (85.04)} & {52.04 (81.16)} & {49.72 (82.28)}\\
    \hline
    {SSBL} & {\textbf{76.72} (\textbf{91.00})} & {\textbf{75.00} ({90.92})} & {\textbf{68.32} (\textbf{87.24})} & {\textbf{64.92} (\textbf{86.32})} & {\textbf{63.80} (\textbf{86.12})} & {\textbf{61.00} (\textbf{86.28})}\\
    \hline
    \hline
    {H2E (200 epochs) \citep{yi2022identifying}} & {78.12 (93.40)} & {74.00 (92.28)} & {69.20 (\textbf{90.44})} & {65.72 (89.04)} & {64.64 (88.56)} & {61.68 (86.88)}\\
    \hline
    {SSBL (200 epochs)} & {{80.48} ({93.16})} & {{76.56} ({92.04})} & {{72.08} ({89.00})} & {{69.52} ({89.04})} & {{66.60} ({88.56})} & {{63.32} ({87.88})}\\
    \hline
    {SSBL+RandAug (200 epochs)} & {\textbf{81.12} (\textbf{93.68})} & {\textbf{76.88} (\textbf{93.72})} & {\textbf{73.08} ({90.16})} & {\textbf{70.16} (\textbf{89.44})} & {\textbf{68.24} (\textbf{88.92})} & {\textbf{65.08} (\textbf{89.68})}\\
    \hline
    \hline
    {ELR+ \citep{liu2020early}} & {78.21 (91.50)} & {73.43 (90.50)} & {55.24 (82.65)} & {52.98 (80.88)} & {51.40 (72.97)} & {50.47 (72.52)}\\
    \hline
    {DivideMix \citep{li2020dividemix}} & {77.68 (\textbf{92.72})} & {75.44 (92.44)} & {61.40 (82.12)} & {62.08 (82.64)} & {54.80 (73.76)} & {54.60 (74.96)}\\
    \hline
    {DivideMix-LA} & {77.44 (91.96)} & {75.96 (91.24)} & {69.80 (\textbf{88.24})} & {68.00 (\textbf{89.32})} & {64.64 (86.44)} & {63.04 (86.92)}\\
    \hline
    {DivideMix-DRW} & {78.28 (92.04)} & {75.44 (\textbf{92.56})} & {65.20 (81.44)} & {65.40 (82.36)} & {57.44 (72.76)} & {57.76 (73.68)}\\
    \hline
    {$\text{SSBL}_2$} & {\textbf{78.56} (91.92)} & {\textbf{76.64} (91.92)} & {\textbf{70.40} ({88.20})} & {\textbf{69.12} ({88.64})} & {\textbf{66.08} (\textbf{87.20})} & {\textbf{63.20} (\textbf{87.40})}\\
    \bottomrule
    \end{tabular}
    }
    
    \caption{Comparison with baselines in Top-1 (Top-5) test accuracy (\%) on original and long-tailed versions of mini WebVision with real-world noise. $\text{SSBL}$ represents our performance of single model, while $\text{SSBL}_2$ represents our ensemble performance of two models.}
    \label{webvision-all}
    \vspace{0.1cm}
\end{table*}

\begin{table*}[t]
\tiny
  \centering
    \resizebox{400pt}{!}{
    \begin{tabular}{l|c |c c c|c c c}
    \toprule
    Dataset  & \multicolumn{7}{c}{CIFAR-100} \\
    \hline
    Noise Rate & \multirow{2}[1]{*}{mean} & \multicolumn{3}{c|}{0.2} & \multicolumn{3}{c}{0.5} \\
\cline{1-1}\cline{3-8}    Imbalance Ratio &       & 10    & 50    & 100   & 10    & 50    & 100 \\
    \hline
    SSBL  &\textbf{48.33} & \textbf{62.45} & \textbf{50.70} & \textbf{43.89} & \textbf{55.65} & 40.82 & \textbf{36.49}\\
    \hline
    w/o rebalancing & 44.68 & 60.98 & 46.55 & 40.19 & 53.05 & 36.41 & 30.91 \\
    \hline
    rebalancing with label frequency & 46.59 & 61.99 & 48.86 & 43.14 & 54.32 & 38.30 & 32.95 \\
    \hline
    w/o $L_{reg}$ in warm up & 47.85 & 61.82 & 49.87 & 43.32 & 54.93 & \textbf{41.02} & 36.16 \\
    \hline
    w/o $L_{reg}$ in the whole process & 36.66 & 61.26 & 14.95 & 12.08 & 55.00 & 40.65 & 36.00 \\
    \hline
    w/o class-aware sample selection & 38.74  & 59.92 & 26.56 & 20.29 & 53.64 & 38.73 & 33.32 \\
    \bottomrule
    \end{tabular}%
    }
  \caption{Ablation study. All experiments are conducted with single model.}
  \label{ablation}%
\end{table*}%

\subsection{Ablation Study}

In order to study the effects of removing components of our method SSBL with single model, we conduct the ablation study on long-tailed versions of CIFAR-100 with symmetric noise and summarize the results in Table \ref{ablation}.

\begin{itemize}
    \item To study the effect of rebalancing the model bias after having estimated the model bias matrix $\bar{M}$, we replace our loss $L'_{\mathcal{L}}$ with the cross-entropy loss $L_{\mathcal{L}}$.
    The decrease in test accuracy suggests that $L'_{\mathcal{L}}$ is effective for learning from long-tailed noisy data.
    \item To study the effect of rebalancing the model bias with the estimated model bias matrix $\bar{M}$ instead of label frequency, we test the performance of rebalancing with label frequency. The degradation in performance verifies the advantage of tackling long-tailed noisy data with the estimated model bias.
    \item To study the effect of the regularization term $L_{reg}$, we test our method without $L_{reg}$ in warm up and without $L_{reg}$ in the whole process. The results imply that $L_{reg}$ in warm up usually brings better performance and $L_{reg}$ in the whole process ensures the robustness under extremely long-tailed distribution.
    \item To study the effect of class-aware sample selection, we replace our class-aware GMM with a single GMM for all classes.
    The degeneration in performance justifies the superiority of class-aware sample selection over sample selection with single GMM. To further demonstrate that class-aware sample selection is robust to long-tailed noisy data, we visualize the precision and recall of samples selected as clean in Figure \ref{pr}. In comparison with sample selection with single GMM, our method with class-aware sample selection achieves comparable or better performance in both precision and recall on all splits of classes.
\end{itemize}

\makeatletter
\newcommand\tabcaption{\def\@captype{table}\caption}
\newcommand\figcaption{\def\@captype{figure}\caption}
\makeatother

\vspace{0.3cm}
\begin{figure}
    \centering
    \begin{minipage}[c]{190pt}
        \centering
        \includegraphics[width=190pt]{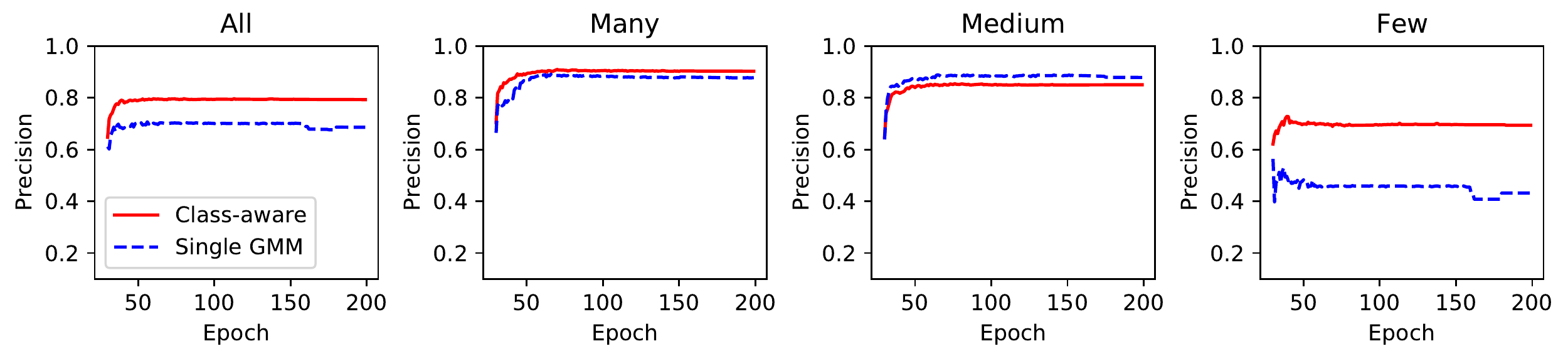}
        \includegraphics[width=190pt]{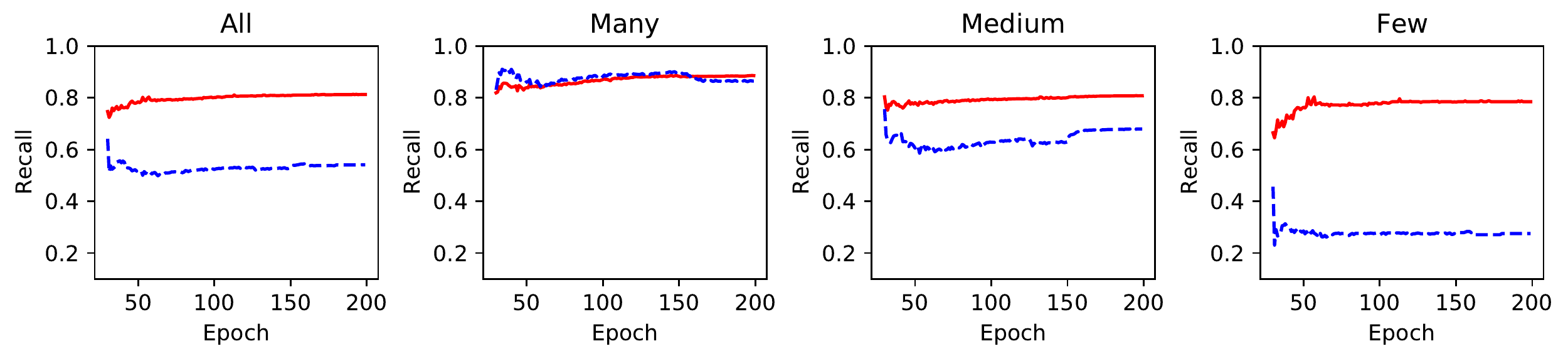}
        \captionsetup{skip=5pt}
        \figcaption{Comparison of precision and recall between class-aware sample selection and sample selection with single GMM. We visualize the precision and recall of samples selected as clean for all and three splits of classes: Many (more than 100 images), Medium (20$\sim$100 images) and Few (less than 20 images). The experiment is conducted on long-tailed CIFAR-100 with symmetric noise. The imbalance ratio $\rho=100$ and the noise rate $r=0.5$, which is an extremely hard setting.}
        \label{pr}
    \end{minipage}
    \hspace{10pt}  
    \begin{minipage}[c]{190pt}
        \centering
        \resizebox{190pt}{!}{%
        \begin{tabular}{c|c|c|c|c|c}
        \toprule
        \multicolumn{2}{c|}{Noise Rate} & \multicolumn{2}{c|}{0.2} & \multicolumn{2}{c}{0.5} \\
        \hline
        \multicolumn{2}{c|}{Imbalance Ratio} & {50} & {100} & {50} & {100} \\
        \hline
        \multirow{2}{*}{$\gamma_{sup} = 3, \gamma_{rel} = 0$} & {Best} & {84.77} & {79.21} & {75.58} & {70.51} \\
        & {Last} & {82.44} & {77.00} & {74.86} & {66.76} \\
        \hline
        \multirow{2}{*}{$\gamma_{sup} = 3, \gamma_{rel} = 0.5$} & {Best} & {85.71} & \textbf{80.73} & {76.48} & {71.93} \\
        & {Last} & {85.01} & \textbf{78.77} & {75.19} & {68.12} \\
        \hline
        \multirow{2}{*}{$\gamma_{sup} = 3, \gamma_{rel} = 1$} & {Best} & \textbf{86.30} & {79.84} & {77.72} & {72.36} \\
        & {Last} & \textbf{85.83} & {78.39} & \textbf{75.76} & {68.44} \\
        \hline
        \multirow{2}{*}{$\gamma_{sup} = 3, \gamma_{rel} = 2$} & {Best} & {79.72} & {73.90} & \textbf{79.08} & {72.79} \\
        & {Last} & {50.70} & {41.86} & {74.54} & {59.28} \\
        \hline
        \multirow{2}{*}{$\gamma_{sup} = 3, \gamma_{rel} = 3$} & {Best} & {79.72} & {73.90} & {78.73} & \textbf{73.22} \\
        & {Last} & {46.29} & {42.03} & {64.68} & {60.22} \\
        \hline
        \multirow{2}{*}{$\gamma_{sup} = 1, \gamma_{rel} = 1$} & {Best} & {85.67} & {80.39} & {75.50} & {69.84} \\
        & {Last} & {85.16} & {78.76} & {73.64} & {67.99} \\
        \hline
        \multirow{2}{*}{$\gamma_{sup} = 2, \gamma_{rel} = 1$} & {Best} & {85.43} & {80.24} & {76.95} & {71.56} \\
        & {Last} & {85.19} & {77.82} & {73.39} & \textbf{68.78} \\
        \hline
        \multirow{2}{*}{$\gamma_{sup} = 4, \gamma_{rel} = 1$} & {Best} & {85.55} & {79.60} & {78.21} & {72.88} \\
        & {Last} & {85.55} & {73.08} & {66.27} & {58.26} \\
        \bottomrule
        \end{tabular}%
        }
        \tabcaption{Ablation study of the hyper-parameters $\gamma_{sup}$ and $\gamma_{rel}$ in test accuracy (\%) on long-tailed CIFAR-10 with symmetric noise.}
        \label{gamma}%
    \end{minipage}
\end{figure}

The hyper-parameters $\gamma_{sup}$ and $\gamma_{rel}$ control the power of suppressing and relaxing, respectively. We conduct experiments on both of them on long-tailed versions of CIFAR-10 with symmetric noise and summarize the results in Table \ref{gamma}. $\gamma_{sup} = 1$ (too little suppression for head classes) or $\gamma_{sup} = 4$ (too much suppression for head classes) will degrade the performance; $\gamma_{rel} = 0$ (without any relaxation for tail classes) or $\gamma_{rel} > 1$ (too much relaxation for tail classes) will harm the performance. In general, $\gamma_{sup} = 3$ and $\gamma_{rel} = 1$ is an appropriate choice.

In our method, we separate training data into a clean labeled set $\mathcal{L}$ and an unlabeled set $\mathcal{U}$ with class-aware sample selection, and then train the model in a semi-supervised manner with our balanced loss. There are other Class-Imbalanced Semi-Supervised Learning (CISSL) methods, e.g., DARP in \citet{kim2020distribution} and ABC in \citet{lee2021abc}. We conduct the ablation study with the existing CISSL methods based on our $\mathcal{L}$ and $\mathcal{U}$, and the results are summarized in Appendix \ref{appendix:cissl} due to limited space. The results show that our method performs better than existing CISSL methods.

\vspace{-0.1cm}
\section{Conclusion}
\vspace{-0.1cm}

We propose the robust method for learning from long-tailed noisy data with sample selection and balanced loss. In specific, we separate the training data into the clean labeled set and the unlabeled set with class-aware sample selection, and then train the model in a semi-supervised manner with a novel balanced loss. Extensive experiments across benchmarks demonstrate that our method is superior to existing state-of-the-art methods.





\bibliographystyle{named}
\bibliography{neurips_2023}


\clearpage
\begin{appendices}

\onecolumn


\section*{\huge Appendix\centering}

\vspace{1cm}

In this appendix, we provide more details about experiments.

\setcounter{secnumdepth}{2}
\renewcommand\thesubsection{\Alph{subsection}}

\subsection{Experiment Setups}
\label{appendix:setup}

\textbf{Datasets.} We conduct experiments on long-tailed and noisy CIFAR-10, CIFAR-100, mini-ImageNet-Red, Clothing1M, Food-101N, Animal-10N and WebVision datasets which are all public. You can find how to generate these datasets in the README.md which is provided in our source code.

\vspace{0.2cm}

\noindent
\textbf{Computing Resources.} We conduct experiments on long-tailed and noisy CIFAR-10, CIFAR-100, mini-ImageNet-Red and Animal-10N datasets with 1 GeForce RTX 2080Ti GPU and conduct experiments on long-tailed and noisy Clothing1M, Food-101N and WebVision datasets with 1 NVIDIA RTX A6000 GPU.

\vspace{0.2cm}

\noindent
\textbf{Implementation Details.} For all datasets, we optimize the model by SGD with a momentum of 0.9 and a weight decay of 0.0005, $\sigma$ is set as 0.9 in EMA, $\gamma_{sup}$ is set as 3 and $\gamma_{rel}$ is set as 1 in $L'_{\mathcal{L}}$. The initial learning rate is reduced by 10 after 50 epochs when the model is trained for 100 epochs, while the initial learning rate is reduced by 10 after 150 epochs when the model is trained for 200 epochs.

On CIFAR-10 and CIFAR-100, the batch size is 64 and the initial learning rate is 0.02. In MixMatch, we set $K=2, T=0.5 ,\alpha=4$ and $\lambda_u \in \{0, 25, 150\}$ following \citet{li2020dividemix} without tuning. For symmetric noise, the warm up period is 10 for CIFAR-10 and 30 for CIFAR-100, the model bias matrix $\bar{M}$ is finished estimating until the 150th epoch and applied in the following training process, we set $\lambda_{warm}=0.2$ and $\lambda_{reg}=0.2$. For asymmetric noise, the warm up period is 140, the model bias matrix $\bar{M}$ is finished estimating until the 170th epoch and applied in the following training process, we set $\lambda_{warm}=0.5$ and $\lambda_{reg}=0.1$.

On mini-ImageNet-Red, the batch size is 64 and the initial learning rate is 0.05. In MixMatch, we set $K=2, T=0.5, \alpha=0.5$ and $\lambda_u=0$. The warm up period is 30, the model bias matrix $\bar{M}$ is finished estimating until the 150th epoch and applied in the following training process, we set $\lambda_{warm}=0.3$ and $\lambda_{reg}=0.2$.

On Clothing1M, the batch size is 32 and the initial learning rate is 0.005. In MixMatch, we set $K=2, T=0.5, \alpha=0.5$ and $\lambda_u=0$. The warm up period is 10, the model bias matrix $\bar{M}$ is finished estimating until the 150th epoch and applied in the following training process, we set $\lambda_{warm}=0.6$ and $\lambda_{reg}=0.1$.

On Food-101N, the batch size is 32 and the initial learning rate is 0.01. In MixMatch, we set $K=2, T=0.5, \alpha=0.5$ and $\lambda_u=0$. The warm up period is 5, the model bias matrix $\bar{M}$ is finished estimating until the 125th epoch and applied in the following training process, we set $\lambda_{warm}=0.6$ and $\lambda_{reg}=0.4$.

On Animal-10N, the batch size is 64 and the initial learning rate is 0.05. In MixMatch, we set $K=2, T=0.5, \alpha=0.5$ and $\lambda_u=0$. The warm up period is 2, the model bias matrix $\bar{M}$ is finished estimating until the 120th epoch and applied in the following training process, we set $\lambda_{warm}=0.6$ and $\lambda_{reg}=0.2$.

On WebVision, the batch size is 32 and the initial learning rate is 0.01. In MixMatch, we set $K=2, T=0.5 ,\alpha=0.5$ and $\lambda_u=0$ following \citet{li2020dividemix} without tuning. The warm up period is 1, the model bias matrix $\bar{M}$ is finished estimating until the 50th epoch (until the 150th epoch when training for 200 epochs) and applied in the following training process, we set $\lambda_{warm}=0.6$ and $\lambda_{reg}=0.4$.

\newpage
\subsection{Comparison with Long-Tailed Learning Methods}
To fully show the effectiveness of our method, we further compare with long-tailed learning methods on long-tailed CIFAR-10 and CIFAR-100 with symmetric noise. Specifically, the selected methods for comparison include LDAM-DRW \citep{cao2019ldam}, Focal \citep{lin2017focal}, CB \citep{cui2019cbloss} and LA \citep{menon2020long}. As shown in Table \ref{appendix}, our method SSBL with single model achieves superior performance over compared long-tailed learning methods.
\begin{table*}[ht]
\Huge

    \centering
    \resizebox{400pt}{!}{%
    \begin{tabular}{c|c|c c c|c c c|c c c|c c c}
    \toprule
    \multicolumn{2}{c|}{Dataset} & \multicolumn{6}{c|}{CIFAR-10} & \multicolumn{6}{c}{CIFAR-100} \\
      \hline
    \multicolumn{2}{c|}{Noise Rate} & \multicolumn{3}{c|}{0.2} & \multicolumn{3}{c|}{0.5} & \multicolumn{3}{c|}{0.2} & \multicolumn{3}{c}{0.5}\\
      \hline
    \multicolumn{2}{c|}{Imbalance Ratio} & 10    & 50    & 100   & 10    & 50    & 100     & 10    & 50    & 100   & 10    & 50    & 100\\
      \hline
    \multirow{2}{*}{ {ERM}} &  {Best} &  {76.90} &  {65.35} &  {60.82} &  {65.75} &  {48.76} &  {39.70} & {45.83} & {35.05} & {29.96} & {28.96} & {19.88} & {16.80}\\
           &  {Last} &  {73.02} &  {61.35} &  {54.48} &  {45.85} &  {33.05} &  {28.79} & {45.64} & {34.93} & {29.88} & {24.33} & {17.77} & {14.47} \\
      \hline
    \multirow{2}{*}{LDAM-DRW \citep{cao2019ldam}} & Best  & 85.23 & 76.90  & 71.75 & 70.07 & 50.36 & 40.92 & 46.96 & 37.54 & 32.28 & 24.68 & 18.79 & 17.26 \\
          & Last  & 85.32 & 77.15 & 72.76 & 70.92 & 53.54 & 44.91 & 47.14 & 37.63 & 32.46 & 28.26 & 19.02 & 17.53 \\
     \hline
    \multirow{2}{*}{Focal \citep{lin2017focal}} & Best  & 73.67 & 61.48 & 53.29 & 45.56 & 34.54 & 28.63 & 45.34 & 33.93 & 30.21 & 25.00    & 16.83 & 15.53\\
          & Last  & 75.99 & 63.78 & 56.76 & 62.79 & 47.94 & 40.39 & 45.55 & 33.95 & 30.21 & 30.13 & 19.17 & 17.05 \\
     \hline
    \multirow{2} {*}{CB \citep{cui2019cbloss}} & Best  & 74.35 & 67.23 & 49.72 & 45.52 & 42.95 & 41.43 & 43.69 & 26.62 & 21.75 & 21.54 & 13.16 & 12.11 \\
          & Last  & 76.52 & 68.52 & 52.22 & 63.76 & 44.66 & 42.09 & 43.91 & 26.74 & 21.96 & 23.19 & 13.80  & 12.34 \\
     \hline
    \multirow{2}{*}{LA \citep{menon2020long}} & Best  & 75.52 & 66.92 & 62.25 & 48.36 & 37.16 & 34.05 & 47.03 & 36.01 & 32.82 & 25.24 & 19.48 & 17.03 \\
          & Last  & 77.84 & 70.88 & 67.55 & 67.89 & 54.61 & 46.94 & 47.21 & 36.09 & 32.93 & 30.78 & 22.84 & 20.98 \\
     \hline
      
    \multirow{2}{*}{SSBL} & Best & \textbf{91.60} & \textbf{86.30} & \textbf{79.84} & \textbf{88.51} & \textbf{77.72} & \textbf{72.36} & \textbf{62.78} & \textbf{51.25} & \textbf{45.68} & \textbf{55.95} & \textbf{42.19} & \textbf{36.87}\\
    & Last &\textbf{91.49} & \textbf{85.83} & \textbf{78.39} & \textbf{88.17} & \textbf{75.76} & \textbf{68.44} & \textbf{62.45} & \textbf{50.70} & \textbf{43.89} & \textbf{55.65} & \textbf{40.82} & \textbf{36.49} \\
    
    \bottomrule
    \end{tabular}%
    }
    \caption{Comparison with long-tailed learning methods in test accuracy (\%) on long-tailed CIFAR-10 and CIFAR-100 with symmetric noise. $\text{SSBL}$ represents our performance of single model.}
  \label{appendix}%
\end{table*}%

\newpage
\subsection{Visualization of Loss Distributions}
\label{appendix:visualize}

\begin{figure}[H]
    \centering
    \subfigure[Noisy vs clean from class 0]{
    \begin{minipage}[t]{0.31\textwidth}
        \centering
        \includegraphics[height=3.7cm]{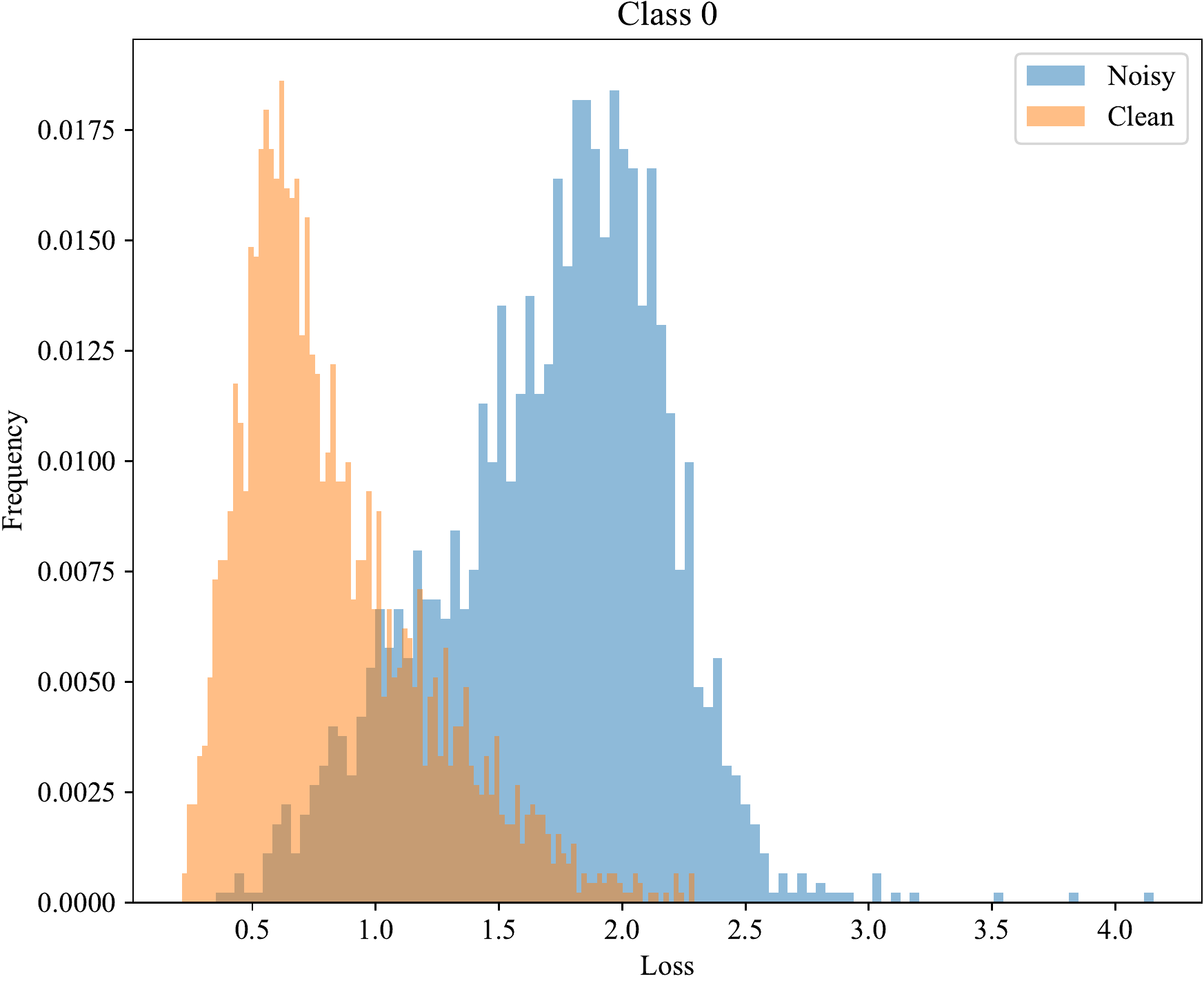}
    \end{minipage}
    }
    \subfigure[Noisy vs clean from class 9]{
    \begin{minipage}[t]{0.31\textwidth}
        \centering
        \includegraphics[height=3.7cm]{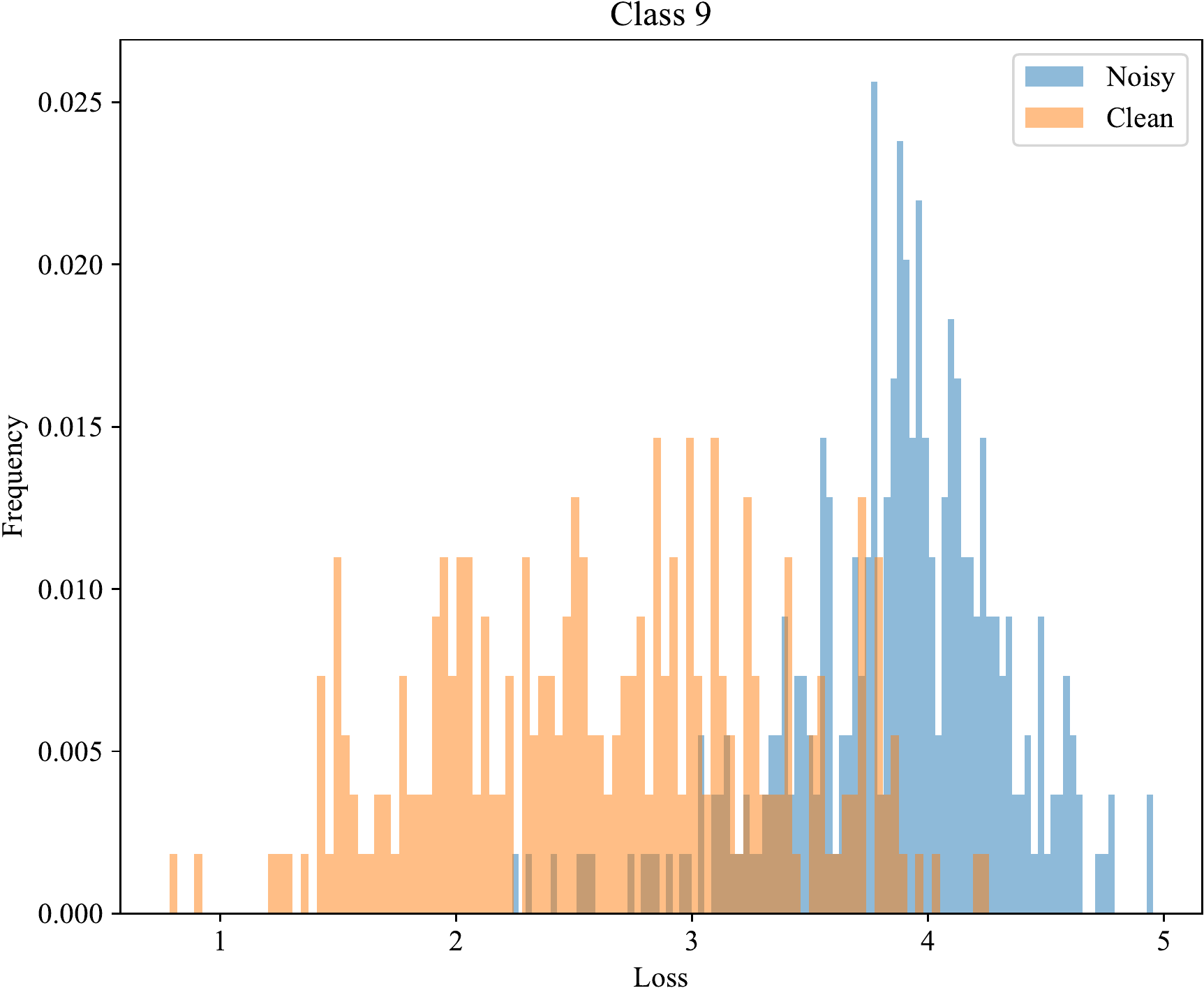}
    \end{minipage}
    }
    \subfigure[Noisy from class 0 vs clean from class 9]{
    \begin{minipage}[t]{0.31\textwidth}
        \centering
        \includegraphics[height=3.7cm]{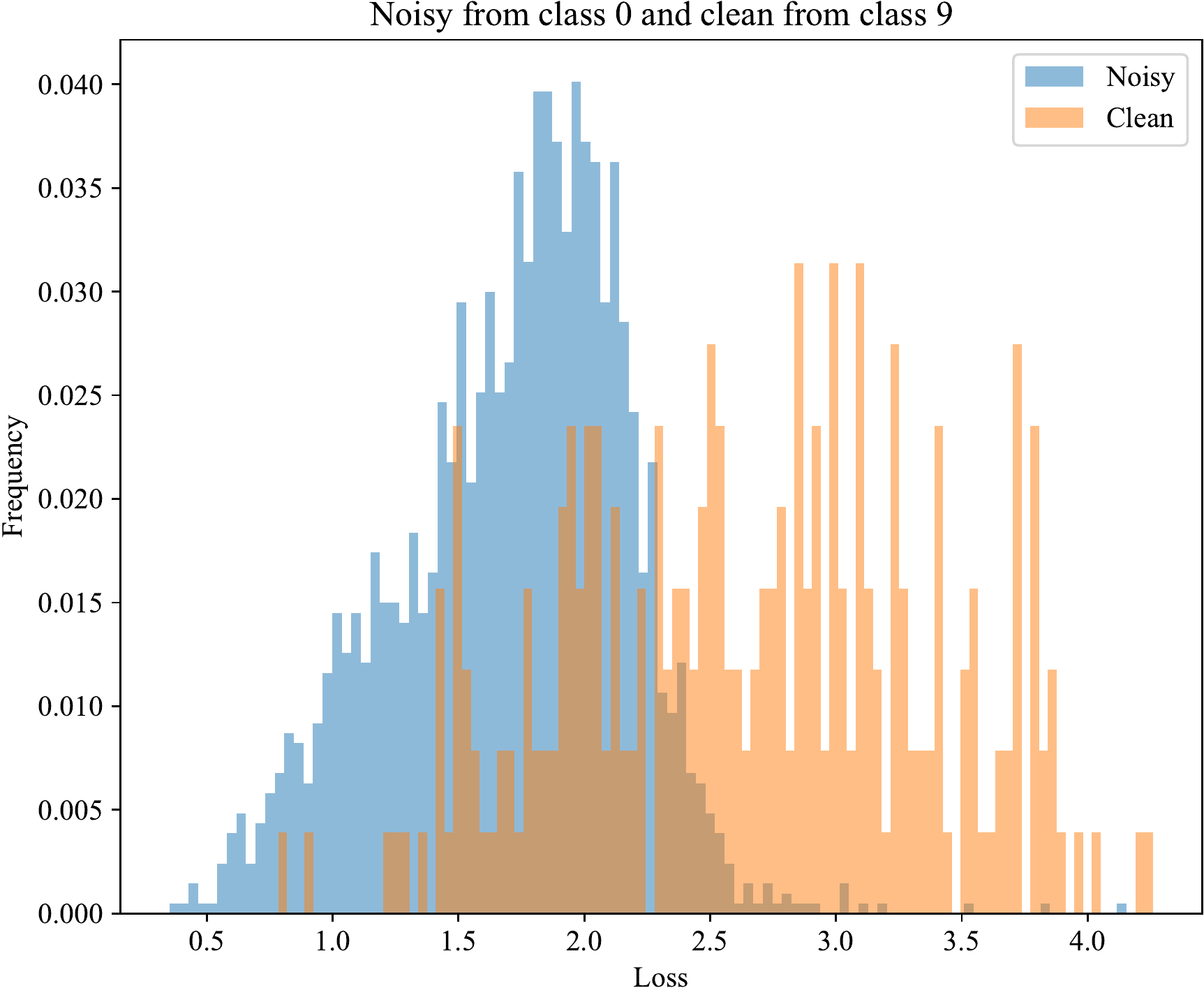}
    \end{minipage}
    }
    \caption{Loss distributions of samples from long-tailed and noisy CIFAR-10 with $r=0.5$ and $\rho=10$. Noisy samples tend to have larger losses than clean samples with the same observed class, while noisy samples from class 0 tend to have smaller losses than clean samples from class 9.} 
    \label{class-level gmm}
\end{figure}

In our method, we propose Class-Aware Sample Selection (CASS) which applies the sample selection by fitting a two-component GMM to the losses of samples class by class. In this section, we further demonstrate the rationality and necessity of the CASS procedure. We train the model with cross-entropy loss on long-tailed CIFAR-10 with symmetric noise for 10 epochs and visualize the loss distributions of samples with the same observed class (class 0 and class 9) in Figure \ref{class-level gmm}(a) and \ref{class-level gmm}(b). We also visualize the loss distribution of noisy samples from class 0 and clean samples from class 9 in Figure \ref{class-level gmm}(c). From the results, we find that: 1) For samples with the same observed class, the noisy samples tend to have larger losses than clean ones, which proves the rationality of small-loss criterion; 2) The losses of samples with different observed classes are not comparable, which justifies the necessity of applying sample selection class by class; 3) In each class, a significant difference exists between the loss distributions of noisy samples and clean samples, which supports the usage of two-component GMM.

\newpage
\subsection{Comparison with Class-Imbalanced Semi-Supervised Learning Methods}
\label{appendix:cissl}

To learn from the labeled clean samples $\mathcal{L}$ and unlabeled samples $\mathcal{U}$ given by the Class-Aware Sample Selection (CASS) procedure, we train the model in a semi-supervised manner with our balanced loss. We combine our CASS with existing class-imbalanced semi-supervised learning methods, i.e., DARP in \citet{kim2020distribution} and ABC in \citet{lee2021abc}, and compare with our method. Specifically, for both DARP and ABC, we change the backbone to PreAct-ResNet-18, keep the total training epochs to 200, warm up the model for 10/30 epochs on CIFAR-10/CIFAR-100 and apply CASS once after our warm up stage. In addition, for DARP, we set its own warm up period as 40$\%$ of the remaining training epochs and keep other hyper-parameters as default. For ABC, we change the initial learning rate to 0.02 and reduce it by a factor 10 after 150 epochs as we do. We report the best and the last test accuracy in Table \ref{semi-long-tailed}. From the results, it can be found that our method SSBL with single model outperforms DARP and ABC under all settings. 

\begin{table*}[ht]
\Huge
    \centering
    \resizebox{400pt}{!}{%
    \begin{tabular}{c|c|c c c|c c c|c c c|c c c}
    \toprule
    \multicolumn{2}{c|}{Dataset} & \multicolumn{6}{c|}{CIFAR-10} & \multicolumn{6}{c}{CIFAR-100} \\
      \hline
    \multicolumn{2}{c|}{Noise Rate} & \multicolumn{3}{c|}{0.2} & \multicolumn{3}{c|}{0.5} & \multicolumn{3}{c|}{0.2} & \multicolumn{3}{c}{0.5}\\
      \hline
    \multicolumn{2}{c|}{Imbalance Ratio} & 10    & 50    & 100   & 10    & 50    & 100     & 10    & 50    & 100   & 10    & 50    & 100\\
    
      \hline
    \multirow{2}{*}{DARP \citep{kim2020distribution}} & Best  & 77.59 & 56.63 & 46.73 & 58.57 & 36.68 & 26.58 & 48.52 & 32.33 & 30.24 & 30.41 & 19.26 & 17.31 \\
          & Last  & 73.61 & 52.33 & 44.33 & 55.53 & 33.94 & 21.22 & 48.49 & 32.25 & 30.24 & 30.41 & 18.75 & 16.15 \\
     \hline
    \multirow{2}{*}{ABC \citep{lee2021abc}} & Best  & 81.25 & 75.53 & 61.43 & 73.63 & 49.35 & 47.12 & 52.88 & 28.37 & 23.39 & 45.15 & 27.53 & 19.61 \\
          & Last  & 76.81 & 57.31 & 45.41 & 63.11 & 25.98 & 36.82 & 46.90  & 20.98 & 11.35 & 38.75 & 22.99 & 19.61 \\
    \hline
    \multirow{2}{*}{ {SSBL}} & Best  & \textbf{91.60}  & \textbf{86.30}  & \textbf{79.84} & \textbf{88.51} & \textbf{77.72} & \textbf{72.36} & \textbf{62.78} & \textbf{51.25} & \textbf{45.68} & \textbf{55.95} & \textbf{42.19} & \textbf{36.87}\\
           & Last  & \textbf{91.49} & \textbf{85.83} & \textbf{78.39} & \textbf{88.17} & \textbf{75.76} & \textbf{68.44} & \textbf{62.45} & \textbf{50.70}  & \textbf{43.89} & \textbf{55.65} & \textbf{40.82} & \textbf{36.49} \\
        \bottomrule
    \end{tabular}%
    }
    \caption{Comparison with class-imbalanced semi-supervised learning methods in test accuracy (\%) on long-tailed CIFAR-10 and CIFAR-100 with symmetric noise. $\text{SSBL}$ represents our performance of single model.}
  \label{semi-long-tailed}%
\end{table*}%

\newpage
\subsection{Model Bias and Label Frequency}
\label{appendix:frequency}

\begin{figure}[H]
    \centering
    \subfigure[$\rho = 10$]{
    \begin{minipage}[t]{0.31\textwidth}
        \centering
        \includegraphics[height=4cm]{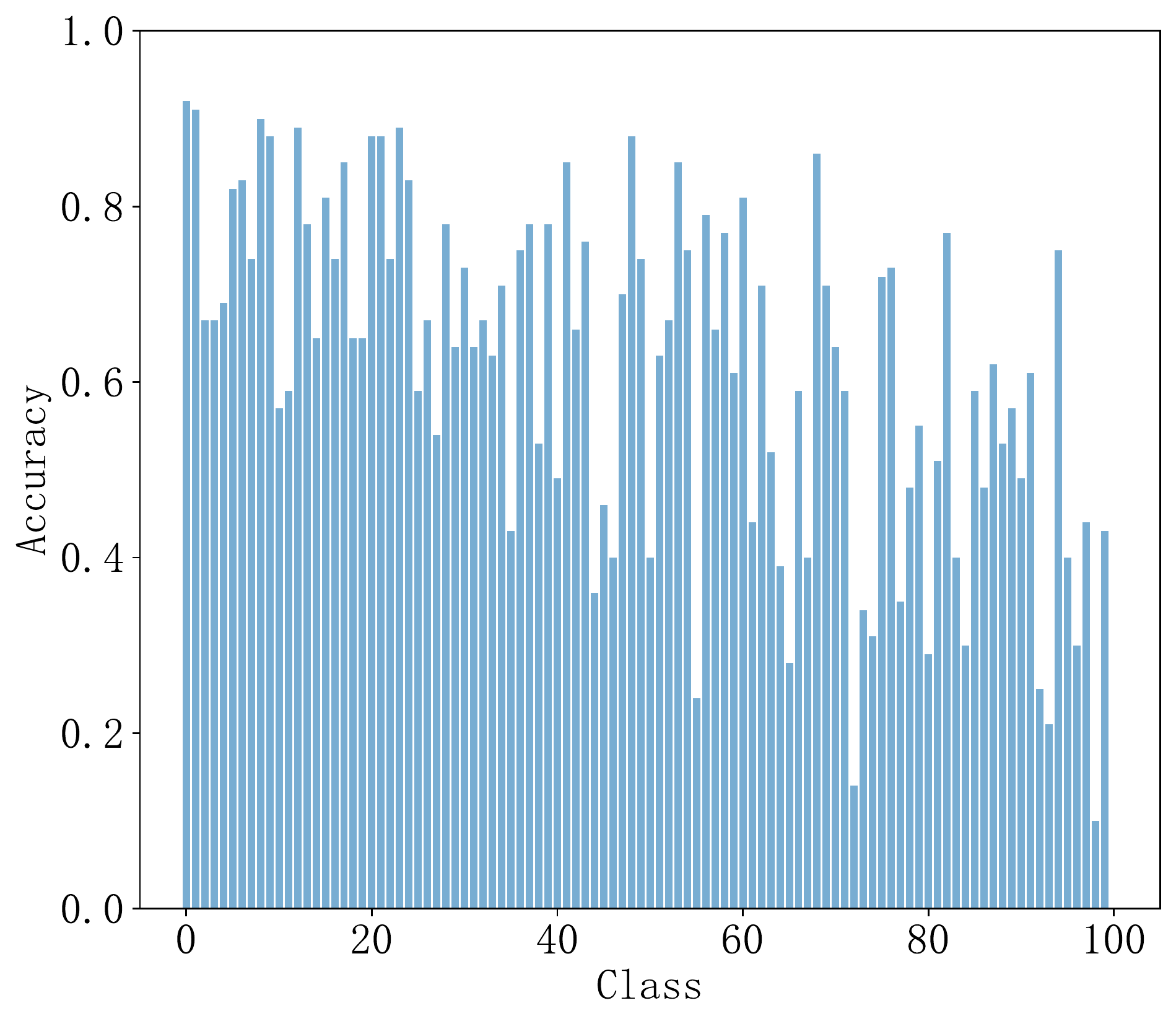}
    \end{minipage}
    }
    \subfigure[$\rho = 50$]{
    \begin{minipage}[t]{0.31\textwidth}
        \centering
        \includegraphics[height=4cm]{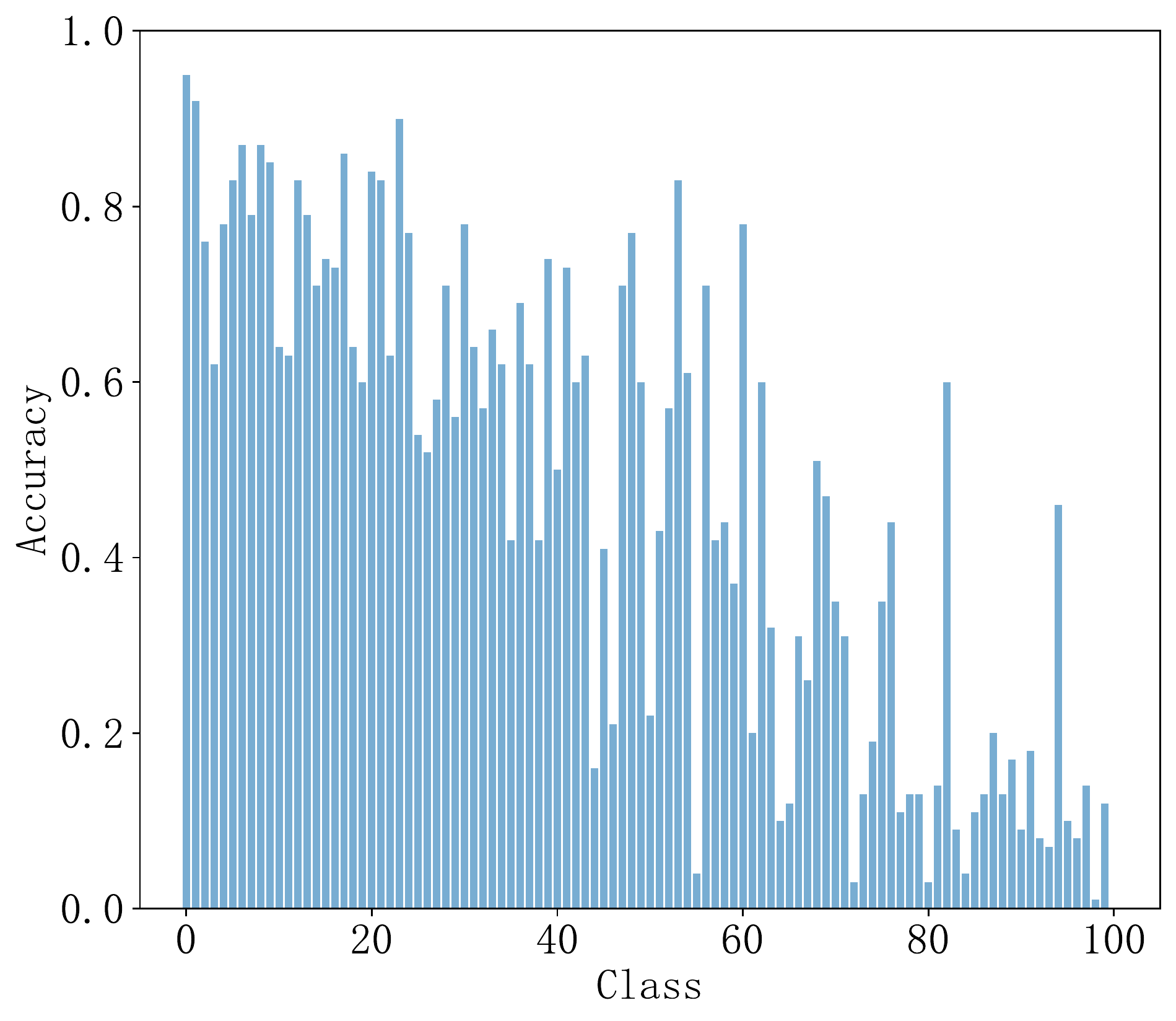}
    \end{minipage}
    }
    \subfigure[$\rho = 100$]{
    \begin{minipage}[t]{0.31\textwidth}
        \centering
        \includegraphics[height=4cm]{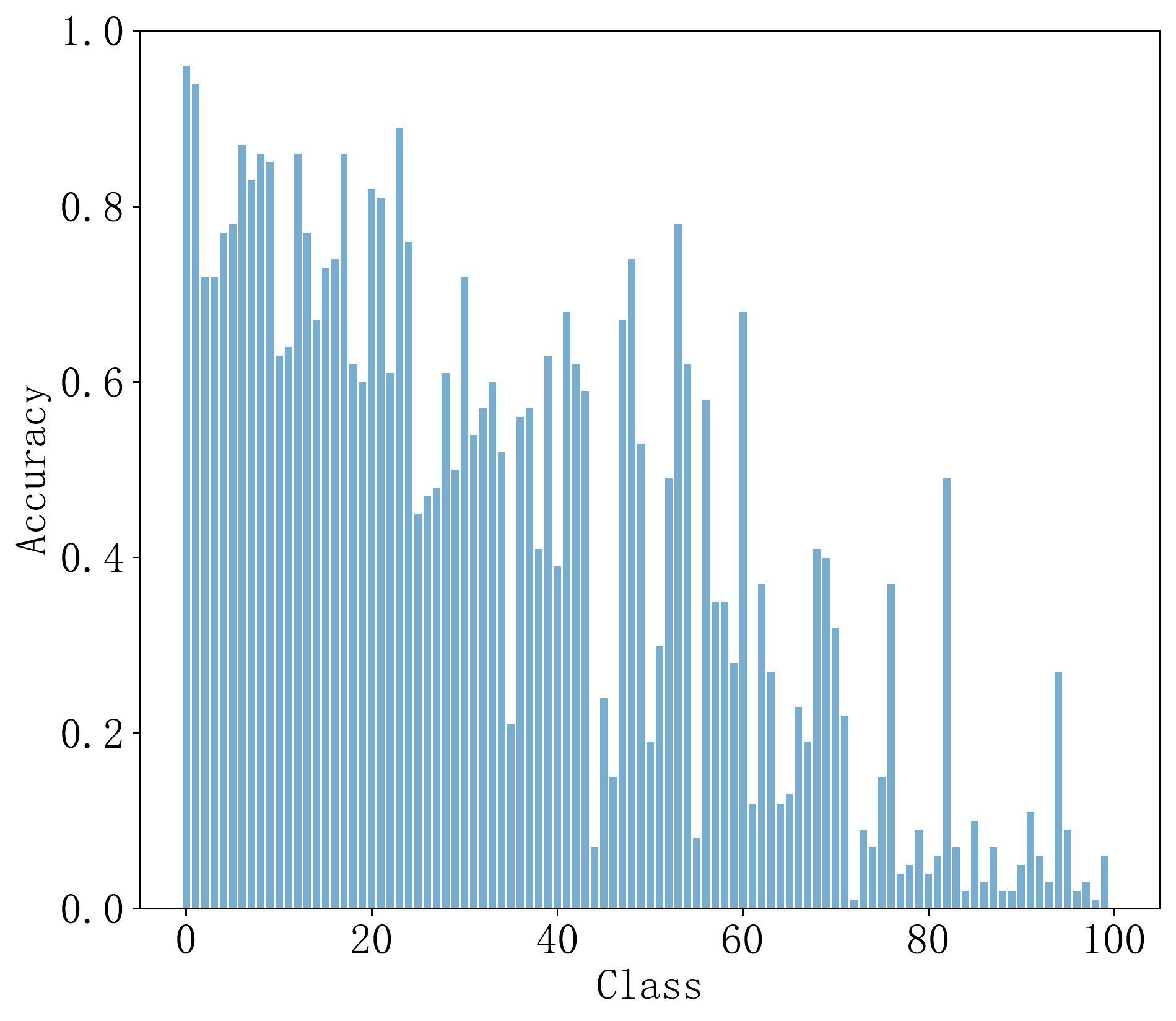}
    \end{minipage}
    }

    \caption{Test accuracy of each class on long-tailed CIFAR-100 with $\rho \in \{10, 50, 100\}$.} 
    \label{frequency and bias}
\end{figure}

To verify that the model bias on different classes may not be directly related to label frequency, we train a PreAct-ResNet-18 with cross-entropy loss for 200 epochs on long-tailed CIFAR-100 with $\rho \in \{10, 50, 100\}$. The results are depicted in Figure \ref{frequency and bias}.
The results show that the test accuracies of some classes with low frequency are higher than that of some classes with high frequency. For example, the test accuracy of class 94 is higher than that of class 35 under all settings.

\end{appendices}

\end{document}